\documentclass[letterpaper]{article} 
\usepackage{algorithm}
\usepackage{amsmath}
\usepackage{tikz}

\usetikzlibrary{positioning, shapes.multipart}
\usepackage{algpseudocode}
\usepackage{aaai2026}  
\usepackage{times}  
\usepackage{helvet}  
\usepackage{courier}  
\usepackage[hyphens]{url}  
\usepackage{graphicx} 
\usepackage{amsfonts} 
\usepackage{bbding}
\usepackage{pifont}
\usepackage{subcaption}
\usepackage{placeins}
\usepackage{bm}  
\usepackage{natbib}  
\usepackage{caption} 
\usepackage{booktabs}
\usepackage{multirow}

\usepackage{newfloat}
\usepackage{listings}
\DeclareCaptionStyle{ruled}{labelfont=normalfont,labelsep=colon,strut=off} 
\floatstyle{ruled}
\newfloat{listing}{tb}{lst}{}
\floatname{listing}{Listing}
\usepackage{tabularx}

\title{Speculative Sampling with Reinforcement Learning}
\author{
    Chenan Wang\textsuperscript{\rm 1},
    Daniel H. Shi\textsuperscript{\rm 1},
    Haipeng Chen\textsuperscript{\rm 1}
}
\affiliations{
    \textsuperscript{\rm 1}William \& Mary

    \{cwang33,dshi01,hchen23\}@wm.edu
}

\usepackage{bibentry}

\begin{document}

\maketitle

\begin{abstract}
Inference time latency has remained an open challenge for real world applications of large language models (LLMs). State-of-the-art (SOTA) speculative sampling (SpS)  methods for LLMs, like EAGLE-3, use tree-based drafting to explore multiple candidate continuations in parallel. However, the hyperparameters controlling the tree structure are static, which limits flexibility and efficiency across diverse contexts and domains.
We introduce \textbf{Re}inforcement learning for \textbf{Sp}eculative \textbf{S}ampling (\textbf{Re-SpS}), the first reinforcement learning (RL)-based framework for draft tree hyperparameter optimization. Re-SpS dynamically adjusts draft tree hyperparameters in real-time, learning context-aware policies that maximize generation speed by balancing speculative aggression with computational overhead. 
It leverages efficient state representations from target model hidden states and introduces multi-step action persistence for better context modeling.
Evaluation results across five diverse benchmarks demonstrate consistent improvements over the SOTA method EAGLE-3,
achieving up to 5.45$\times$ speedup over the backbone LLM and up to 1.12$\times$ speedup compared to EAGLE-3 across five diverse benchmarks, with no loss in output fidelity.
\end{abstract}

\begin{links}
    \link{Code}{https://github.com/wmd3i/ReSpS.git}
\end{links}

\section{Introduction}
Inference time latency has been an open challenge in deploying large language models (LLMs) \cite{openai2023gpt4,touvron2023LLaMA,claude3, dubey2024LLaMA}. This latency, which stems from the sequential nature of autoregressive decoding and large model sizes, severely restricts the viability of LLMs in many time-sensitive applications \cite{kaplan2020scaling, li2024large, urlana2024llms}. To address this, speculative sampling (SpS) techniques aim to accelerate generation by parallelizing the process without compromising correctness, thereby enabling more efficient deployment of these massive models \cite{chen2023accelerating, leviathan2023fast, sun2023spectr, jeon2024recursive, miao2024specinfer}.

\begin{figure}[t]  
    \centering
    \includegraphics[width=0.45\textwidth]{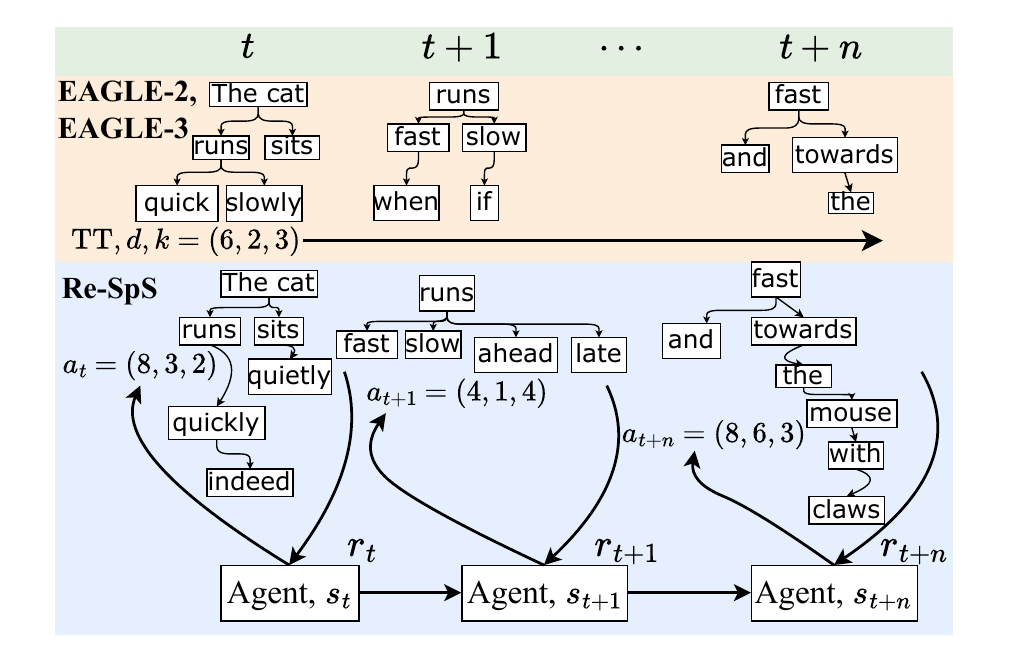}
    \caption{Re-SpS vs. EAGLE-2~\&~3: Comparison of speculative sampling tree structures through steps $1\cdots n$ of a generation task.
    \textbf{Top: }EAGLE-2~\&~3 uses static hyperparameters (upper limits for the total token $TT$, depth $d$, and expansion factor $k$).
    \textbf{Bottom: }Re-SpS dynamically adapts the draft tree hyperparameters  based on the context. Step $t+1$ exemplifies a more cautious draft tree, tuning top-k up and depth down to safely capture more possible tokens without risking costly rejections. Step $t+n$ shows a more aggressive configuration, with depth adjusted to allow for confident drafting. Draft tokens and branching structures were manually chosen to provide a readable, illustrative comparison. Actual model outputs may vary, but the figure faithfully reflects the parameter and adaptivity constraints of each approach. }
    \label{fig:resps_tree_comp}
\end{figure}

SpS  accelerates language model inference by first using a lightweight draft model to generate candidate tokens, which are then jointly verified by the larger target model in a single pass. If any candidate is rejected, the process resumes from the last accepted token; if all are accepted, the target model generates the next token and the cycle repeats. This approach reduces the number of full model forward passes,  enabling faster generation without sacrificing accuracy. This fundamental concept has been extended, first by SpecInfer \cite{miao2024specinfer} and subsequently by other methods, which replace the linear chain of candidates with a ``draft tree," enabling the exploration of multiple potential continuations simultaneously \cite{cai2024medusa,li2024eagle2,li2025eagle3, wang2025opt, zheng2025faster}.
For example, EAGLE-2 \cite{li2024eagle2} and EAGLE-3 \cite{li2025eagle3}, the SOTA methods, adopt dynamic draft tree shaping, with EAGLE-3 taking a leap forward by recurrently feeding the draft model's unverified predictions as input for making further predictions, alongside feature sequences of the target model's hidden layers. Figure \ref{fig:resps_tree_comp} (upper half) illustrates the sampling procedure of EAGLE-2 and EAGLE-3.
While these works allow the nodes in the draft tree to take on different orientations, the hyperparameters governing overall structure—such as depth and branching factor—remain fixed and hand-tuned.

Since reinforcement learning (RL) \cite{sutton1998reinforcement,schulman2017proximal} is a natural fit for sequential optimization problems, we take an RL-based approach that can adjust SpS draft tree hyperparameters dynamically. 
While promising, the actual implementation of a vanilla RL algorithm poses a central challenge: frequent RL policy calls incur substantial computational overhead that is comparable with the computation gains of improved SpS.

We identify two distinct sources of this computational overhead. First, \textit{the state representation overhead}: generating rich, context-sensitive state embeddings at every decoding step---such as SentenceBERT \cite{reimers-gurevych-2019-sentence}---incurs significant latency. In many cases, this encoding cost (often several milliseconds per step) can exceed the time saved by speculative sampling itself, especially when applied to long sequences where embedding computation is invoked repeatedly. Second, \textit{the policy inference overhead}: each adaptation of the speculative sampling hyperparameters requires a forward pass through the RL policy network. When policy inference is performed at every generation step, the cumulative runtime cost quickly adds up over long sequences, creating a fundamental trade-off: while more frequent adaptation enables more context-aware and potentially optimal draft tree control, it also results in prohibitively high computational expense.

In this paper, we propose \textbf{Re}inforcement learning for \textbf{Sp}eculative \textbf{S}ampling (\textbf{Re-SpS}), the first framework to formulate the control of SpS draft tree hyperparameters as a reinforcement learning problem. Re-SpS introduces two key innovations to address computational overhead: (1) an \textit{efficient feature reuse} mechanism, which leverages the target model's internal hidden states (specifically, a concatenated multi-layer feature vector from the draft model) as rich, low-cost state representations—eliminating the need for a separate encoder; and (2) a \textit{multi-step action persistence} strategy, where a selected hyperparameter configuration is cached and reused across multiple decoding steps, thereby amortizing the cost of policy inference. As illustrated in Figure~\ref{fig:resps_tree_comp}, this enables the Re-SpS policy to observe the generation context and dynamically select draft tree upper limits for the total tokens, depth, and expansion factor at each draft-verify run, optimizing tree-based SpS hyperparameters more efficiently than the static approaches.

Our key contributions include: 1) We propose Re-SpS, the first RL-based framework for SpS draft tree hyperparameter optimization. 
2) We identify sources of the key technical challenge of computational overhead from RL policy calls, and introduce two technical innovations—efficient feature reuse and multi-step action persistence—to address them.
3) We conduct extensive empirical evaluations across five diverse benchmarks and three model sizes, demonstrating consistent speedup improvements (up to 5.45$\times$ over the backbone LLM and up to 1.12$\times$ over the SOTA method EAGLE-3) while maintaining exact output fidelity.
Ablation studies further validate the effectiveness of our technical designs.

\section{Related Work}%

This section outlines two methodological themes for SpS.\smallskip

\noindent\textbf{Chain and Tree-based Verification.}
The early SpS methods focus on a linear chain-based draft sequence \cite{chen2023accelerating, leviathan2023fast}. Recent works, starting with SpecInfer \cite{miao2024specinfer}, have leveraged tree-based structures to systematically explore and optimize token candidate selection \cite{zhang2024survey}.
Many architectural innovations have emerged to optimize tree-based speculation \cite{sun2023spectr,he2024rest,  cai2024medusa, li2024eagle, jeon2024recursive, chen2024sequoia}. SpecInfer \cite{miao2024specinfer} introduced parallel token verification using tree attention \cite{vaswani2017attention} and fixed expansion hyperparameters. Medusa \cite{cai2024medusa} integrated drafting into the target model via MLP heads, while EAGLE \cite{li2024eagle} combined tree-based drafting with feature-level autoregression. C2T \cite{huo2025c2t} employed lightweight classifiers for tree management.
However, all these methods share a common limitation: they use static hyperparameters for draft tree generation.

\noindent\textbf{Adaptive Control of Draft Tree Structures.}
A major thrust of recent research has been toward heuristic-driven adaptivity \cite{zhang2024survey}. EAGLE-2 \cite{li2024eagle2} introduces a context-aware dynamic draft tree, which uses the confidence of the draft model as a proxy for the final acceptance rate to intelligently prune the unlikely branches of the speculative tree and re-rank all nodes to select the most promising candidates for verification \cite{li2024eagle2}. The most recent iteration, EAGLE-3, abandons feature prediction for direct token prediction and incorporates target model hidden states as well as past draft outputs into draft model inputs, but retains static draft tree hyperparameters \cite{li2025eagle3}. Complementary adaptive strategies have also gained traction, such as SpecDec++ \cite{huang2024specdec++}, DySpec \cite{xiong2025dyspec}, OPT-Tree \cite{wang2025opt}, and ProPD \cite{zhong2024propd}. The latter two enlist early pruning and dynamic tree generation strategies--however, these approaches remain limited: OPT-Tree optimizes node allocation within fixed budgets, while ProPD uses regression-based tree sizing. There have been some initial explorations on data-driven and learning-based frameworks for SpS hyperparameter optimization. For example,  HASS uses distillation-based supervised learning to better align the draft model's predictions with the target's \cite{zhanglearning}, and MetaSD and BanditSpec use multi-armed bandit algorithms to select drafters (MetaSD) or distinct speculative sampling strategies \cite{kim2024unified, hou2025banditspec}. While these methods focus on selecting between pre-defined configurations or models, none explore learning adaptive policies that can dynamically adjust draft tree structure hyperparameters based on generation context.

\section{Preliminary}

\subsection{Auto-Regressive Decoding in LLMs}

Large Language Models (LLMs) generate text one token at a time using auto-regressive decoding \cite{bengio2003neural}. Given a discrete token sequence $x = (x_1, x_2, \dots, x_s) \in \mathcal{T}^s$ of length $s$ over a token set $\mathcal{T}$, we define a slice of this sequence at decoding round $t$ as $x^t_{1:m} = (x_1, x_2, \dots, x_m)$. The LLM outputs a probability distribution over the next token conditioned on all previous tokens. Specifically, the probability of token $x_t$ is given by $P_{\text{LLM}}(x_t \mid x_1, \dots, x_{t-1})$.

To generate the next token, a sampling method (e.g., greedy, top-$k$ sampling~\cite{fan2018hierarchical}) is applied to this distribution. We define $x_0$ as the prompt or query tokens and let the model generate an output sequence of $m$ tokens, denoted by $y_1, \dots, y_m$. In this work, we use greedy decoding for reproducibility, where each token is deterministically chosen as:
\[
y_i = \arg\max_y P_{\text{LLM}}(y \mid x_0, y_1, \dots, y_{i-1}), \quad i = 1, \dots, m.
\] 

\subsection{Speculative Sampling}

Speculative Sampling \cite{leviathan2023fast, chen2023accelerating} introduces a \emph{draft-then-verification} paradigm to accelerate auto-regressive decoding. We formalize this process mathematically to establish notation for subsequent analysis.

\textit{Draft Phase.} At decoding step $t$, given context $\mathbf{c}_t = (x_0, y_1, \dots, y_{t-1})$ consisting of prompt $x_0$ and accepted tokens, a draft model $M_d$ generates a candidate sequence $\mathbf{\hat{y}}_t = (\hat{y}_{t}, \hat{y}_{t+1}, \dots, \hat{y}_{t+n-1})$ of length $n$:
\[
\mathbf{\hat{y}}_t \sim M_d(\cdot \mid \mathbf{c}_t)
\]

\textit{Verification Phase.} The target model evaluates the extended sequence $(\mathbf{c}_t, \mathbf{\hat{y}}_t)$ in a single forward pass, yielding conditional distributions:
\[
P_T(y_{t+i} \mid \mathbf{c}_t, \hat{y}_t, \dots, \hat{y}_{t+i-1}), \quad i = 0, \dots, n-1
\]

\textit{Acceptance Criterion.} Each draft token $\hat{y}_{t+i}$ is accepted with probability determined by the acceptance function $\alpha$:
\[\alpha(\hat{y}_{t+i}) = \min\left(1, \frac{P_T(\hat{y}_{t+i} \mid \mathbf{c}_t, \hat{y}_t, \ldots, \hat{y}_{t+i-1})}{P_d(\hat{y}_{t+i} \mid \mathbf{c}_t, \hat{y}_t, \ldots, \hat{y}_{t+i-1})}\right)\]
where $P_T$ and $P_d$ represent the probability distributions from the target model and draft model, respectively
The acceptance length $\ell_t$ is defined as the number of consecutively accepted tokens:
\[
\ell_t = \min\{j \in \{0, 1, \dots, n\} : \alpha(\hat{y}_{t+j}) = 0\} \cup \{n\}
\]

\textit{Draft Tree Expansion.}
EAGLE-2 \cite{li2024eagle2} introduces a two-phase dynamic construction process that moves beyond EAGLE's \cite{li2024eagle} static draft tree approach. Unlike EAGLE's fixed tree structure, EAGLE-2 dynamically adjusts the draft tree based on context-dependent acceptance rates rather than assuming acceptance depends solely on token position. EAGLE-2 constructs the tree through selective layer-wise expansion. At each layer, the algorithm selects the top-k tokens with highest global acceptance probabilities from the entire current layer for expansion to the next layer. The selection is based on global acceptance probability $V_i$, calculated as the cumulative confidence score along the path from root to each token:
\[
V_i = \prod_{i=1}^{m} c(y_i \mid y_1, \dots, y_{i-1})
\]
where $c(\cdot)$ represents the draft model's confidence in token acceptance. 

\textit{Confidence-Based Candidate Reranking.} EAGLE-2 \cite{li2024eagle2}  introduces reranking based on $V_i$: after draft tree expansion is complete, \textit{all} tokens in the tree are reranked before being used as candidates for verification.

\begin{figure*}[t]  
    \centering
    \includegraphics[width=0.9\textwidth]{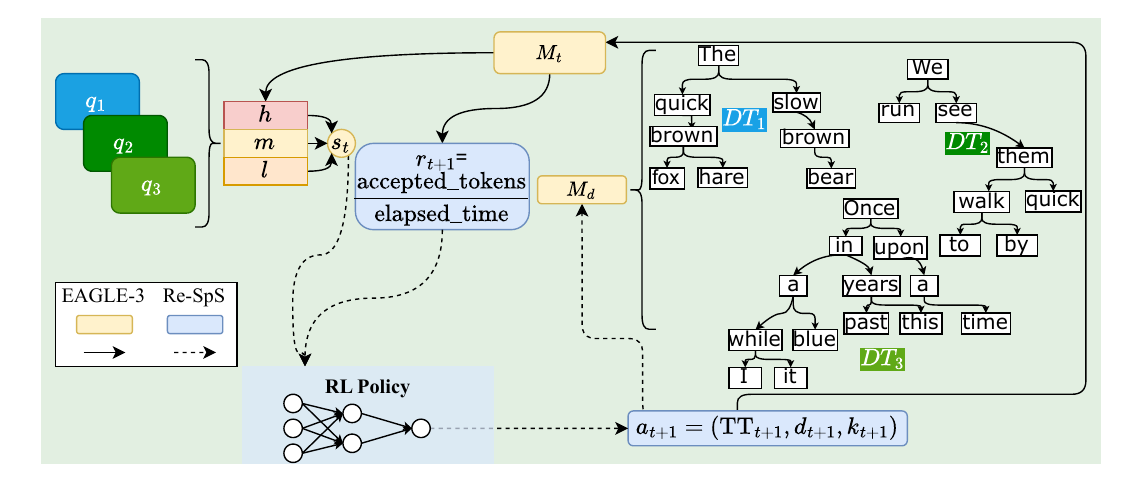}   
    \caption{Architecture of Re-SpS: The diagram illustrates the Re-SpS framework for SpS in LLMs. For each new input prefix (from question tasks $q_1, q_2, q_3, \ldots$), hidden state vector ($h, m, l$) from the target model ($M_t$) is aggregated into $s_t$ and passed to a reinforcement learning (RL) policy. The RL agent outputs draft tree hyperparameters $(TT_{t+1}, d_{t+1}, k_{t+1})$ for the next generation step. The draft model ($M_d$) constructs tree-structured speculative candidates ($DT_1, DT_2, DT_3$), which are verified by the target model. The number of accepted tokens  and elapsed time  are recorded, and the generation speed \(r_{t+1} = \frac{\text{accepted tokens}}{\text{{elapsed time}}}\) is used as the RL reward. Solid arrows show EAGLE-3's static pipeline; dashed arrows highlight Re-SpS's adaptive, RL-driven control. Note: For clarity, the figure is simplified; in practice, the draft trees $DT$ are redefined multiple times for each new prefix within a question task until a maximum sequence length or end-of-sequence is reached.}
    
    \label{fig:resps_arch}
\end{figure*}

\section{Methodology}
Despite notable progress from recent tree-based SpS methods such as EAGLE-2 and EAGLE-3~\cite{li2024eagle2,li2025eagle3}, a critical limitation remains: these approaches rely on static draft tree hyperparameters throughout the decoding process. As a result, their efficiency and adaptability are inherently restricted, especially when handling diverse or dynamically changing contexts.
To overcome these obstacles, we introduce Re-SpS, the first RL-based approach for draft tree hyperparameter optimization, alongside innovative designs that address the computational overhead of RL policy calls.

\subsection{MDP Formulation and Vanilla RL Solution}
We first formulate the problem of dynamically selecting speculative sampling hyperparameters as a Markov Decision Process (MDP), defined by the tuple $(\mathcal{S}, \mathcal{A}, R, \xi)$.

\paragraph{State Space $\mathcal{S}$} At each decision point $t$, the agent is in a given state $s_t \in \mathcal{S}$ that captures the current context of the text generation process. This state should contain sufficient information for the agent to make an informed decision about the optimal speculative strategy. A naive approach would be to use a rich embedding of the entire context (prompt, question, and all previously generated tokens) using a powerful model like SentenceBERT~\cite{reimers-gurevych-2019-sentence}:
\begin{equation}
s_t = [e(\text{prompt}, y_{<t})],
\label{eq:state_sbert}
\end{equation}
\noindent where $e(\cdot)$ denotes the embedding function. 

\paragraph{Action Space $\mathcal{A}$} The action space consists of discrete hyperparameter combinations drawn from predefined sets of values:
\begin{equation}
\mathcal{A} = \left\{ (\text{TT}, d, k) \,\middle|\,
\begin{aligned}
  & \text{TT} \in \mathcal{S}_{\text{TT}} \\
  & d \in \mathcal{S}_{d} \\
  & k \in \mathcal{S}_{k}
\end{aligned}
\right\}
\label{eq:action_space}
\end{equation}
\noindent where $\texttt{TT}$, $d$, and $k$ represent the upper limits for the total number of tokens, tree depth, and top-$k$ expansion factor per layer, respectively. The sets $\mathcal{S}_{\textit{TT}}$, $\mathcal{S}{d}$, and $\mathcal{S}_{k}$ are finite collections of pre-selected integer values for each hyperparameter. This action space allows the agent to select a specific draft tree configuration at each decision point, enabling dynamic adaptation to the current context (see Figure~\ref{fig:resps_arch}).

\paragraph{Reward Function $R$} Upon executing an action, the agent receives a reward $r_t = R(s_t, a_t)$. We define the immediate reward as the generation speed, measured in accepted tokens per second calculated by dividing the number of accepted tokens by the elapsed time for the drafting/verification step, which directly aligns the agent's objective with our goal of minimizing latency.
\begin{equation}
r_t = \frac{\text{accepted tokens}}{\text{elapsed time (seconds)}}
\label{eq:reward_immediate}
\end{equation}
This reward function captures the efficiency of the speculative sampling process, incentivizing the agent to select hyperparameters that maximize the number of accepted tokens while minimizing the time taken for generation.

\paragraph{Transition Function $\bm{\xi}$} The transition function ${\xi}(s_{t+1} \mid s_t, a_t)$ defines how the state evolves based on the selected action. In our case, it is implicitly and deterministically defined by the draft tree construction process, which generates a new state based on the current state (i.e., feature vector) and selected hyperparameters, and by the speculative decoding process itself, where the state evolves as new tokens are generated and accepted.

\paragraph{Vanilla RL Implementation}
We use PPO~\cite{schulman2017proximal} as our backbone reinforcement learning algorithm due to its stability and effectiveness in sequential decision-making tasks. PPO updates the policy by maximizing the following clipped surrogate objective:
\begin{equation*}
L^{\mathrm{PPO}}(\theta) = \mathbb{E}_t \left[ 
  \min\left( r_t(\theta) \hat{A}_t,\, \mathrm{clip}\big(r_t(\theta), 1-\epsilon, 1+\epsilon\big) \hat{A}_t \right) 
\right]
\end{equation*}
where $r_t(\theta) = \frac{\pi_\theta(a_t|s_t)}{\pi_{\theta_\mathrm{old}}(a_t|s_t)}$ is the probability ratio and $\hat{A}_t$ is the estimated advantage.

We also explore the maximum entropy variant, which augments the standard PPO objective with an entropy regularization term. This addition encourages the policy to maintain high exploration by favoring more stochastic action distributions, thereby mitigating premature convergence to suboptimal policies. The overall objective is given by:
\[
L^{\mathrm{MAX\text{-}ENT}}(\theta) = L^{\mathrm{PPO}}(\theta) + \beta_H \, \mathbb{E}_t \big[ H(\pi_\theta(\cdot|s_t)) \big]
\]
where $H(\pi_\theta(\cdot|s_t))$ is the entropy of the policy at state $s_t$, and $\beta_H$ is a weight factor~\cite{haarnoja2018soft,schulman2017proximal}.

\paragraph{Lossless Output Fidelity}
Re-SpS is built upon EAGLE-3~\cite{li2025eagle3} and inherits its lossless property: it uses SpS with target-model verification~\cite{leviathan2023fast,chen2023accelerating}, preserving the output distribution of standard autoregressive decoding. In contrast, methods like MEDUSA~\cite{cai2024medusa} relax the acceptance conditions of SpS and lack rejection-based correction, thereby failing to guarantee distributional equivalence under non-greedy decoding~\cite{li2025eagle3}.

\subsection{Motivation: Challenges in a Naive RL Implementation}
While the MDP formulation is straightforward, a naive implementation RL implementation (e.g., based on vanilla PPO or Max-Entropy PPO), where the agent makes a decision at every single decoding step, presents significant practical challenges. Our initial explorations revealed that this approach fails because the computational overhead introduced by the RL agent can easily outweigh the latency gains from improved speculation. This overhead stems from two primary sources.

First, generating a rich state representation for the RL agent is prohibitively expensive. A robust policy requires a detailed understanding of the current generation context. A naive approach would be to encode the entire context (prompt, question, and all previously generated tokens) at each step using a powerful model like SentenceBERT \cite{reimers-gurevych-2019-sentence} via Equation \ref{eq:state_sbert}. However, this would introduce significant inference latency (approximately 5--15 ms per call) and memory overhead even with a small model, likely negating any speedup from the speculative decoding itself, rendering the entire process slower than the baseline it is meant to accelerate. In addition to the computational latency, maintaining and processing 384-dimensional embedding vectors for multiple contexts also increases the memory footprint considerably. Especially when decisions may be needed every decoding step, this overhead accumulates rapidly, potentially negating any speedup gained through speculative decoding.

The second major source of overhead arises from the frequency of RL policy network inference. Making hyperparameter decisions at every decoding step enables fine-grained, highly adaptive control via Equation \ref{eq:reward_immediate} over the draft tree structure, but it also creates prohibitive computational costs. Specifically, each policy call requires a forward pass through the neural network, which typically consists of 2--4 layers with 64--512 hidden units. While more frequent decisions can provide better adaptation to the evolving context, this increased decision frequency also raises the inference overhead proportionally. Over the course of generating a single response—which can involve 50--100 or more decoding steps per turn and 2--5 turns per question—the cumulative cost of repeated policy calls can quickly outweigh any computational savings achieved through a more optimized draft tree or improved speculative sampling. Thus, the compounded inference time from frequent policy queries may become a major bottleneck, ultimately negating the efficiency benefits of the approach.

Together, these factors make it likely that this method would render the overall process slower than the baseline it is intended to accelerate.

\subsection{Addressing the Challenges with Re-SpS}
Our Re-SpS framework is designed to solve this MDP while explicitly addressing the computational overheads identified above through two key innovations.

\paragraph{Efficient State Representation via Feature Reuse}
To resolve the challenge of state representation overhead, we redefine the state $s_t$ not with a costly external embedding, but by reusing the internal features already computed by the EAGLE-3 \cite{li2025eagle3} draft model. Specifically, the state is a concatenation of hidden states from three strategically selected layers of the target language model:
\begin{equation}
s_t = [h^{(h,m,l)}_{\text{LM}}]
\label{eq:state_representation}
\end{equation}
where $h^{(h,m,l)}_{\text{LM}}$ is the fused feature vector, shown as $s_t$ in Figure \ref{fig:resps_arch}. This approach provides a rich, multi-level representation of the generation context—capturing syntactic, semantic, and task-specific information—without introducing any additional inference cost, as these features are an integral part of the EAGLE-3 architecture.
In EAGLE-3, these features serve as hidden states for the draft model, enabling it to generate draft tokens; similarly, we utilize them to construct the state representation for our RL agent. The distinction is that, while EAGLE-3 passes these hidden states through a fully connected layer to obtain a single fused feature vector, our approach instead concatenates the hidden states from the three layers directly, thereby reducing the computational cost.
\paragraph{Amortizing Policy Inference with Multi-Step Action Persistence}
To mitigate the overhead of frequent policy calls, we introduce a multi-step action persistence mechanism, or action caching. Instead of querying the policy network at every step, a selected action $(TT, d, k)$ is cached and reused for $N$ consecutive decoding steps---that is, the \textit{cache interval} has length $N$ ($N=10$ during training and $N=30$ during inference). This cache interval amortizes the cost of a single policy inference over multiple decoding steps, striking a balance between adaptivity and efficiency. This approach leverages the Markov property, as the reward signal, which we average over the $N$ decoding steps in the cache interval, naturally captures the temporal dynamics and performance impact of the chosen action without requiring a complex, multi-step state history. To complement this, we compute the reward as an average across the cache duration:
\begin{equation}
r_{\text{avg}} = \frac{1}{N} \sum_{i=1}^{N} \frac{\text{accepted\_tokens}_i}{\text{elapsed\_time}_i \text{(seconds)}}
\label{eq:reward_computation}
\end{equation}
This allows that the reward signal reflects the cumulative performance impact of the cached hyperparameter decision. The core logic of this process is detailed in Algorithm~\ref{alg:resps_action_caching}. The overall architecture of Re-SpS is illustrated in Figure~\ref{fig:resps_arch} and the overall algorithm in Algorithm~2 in the Appendix~B.
The algorithm is designed to be efficient and adaptive to all tree-based speculative sampling methods, such as Medusa \cite{cai2024medusa}, EAGLE-2 \cite{li2024eagle2}, and EAGLE-3 \cite{li2025eagle3}.

\begin{algorithm}[htb!]
\caption{Re-SpS with Action Caching}
\label{alg:resps_action_caching}
\begin{algorithmic}[1]
\Require Target model $M_t$, draft model $M_d$, policy $\pi_\theta$, cache interval length $N$, cache step $c\in [0, N]$
\Ensure Draft tree $DT$
\State $c \gets 0$ \Comment{Initialize cache step within interval}
\State $s_t \gets \text{concat}(h, m, l)$ from $M_t$
\If{$cache\_step = 0$ or no cached action}
    \State $(TT, d, k) \gets \pi_\theta(s_t)$
    \State $cache\_step \gets 1$
\Else
    \State Use cached $(TT, d, k)$
    \State $cache\_step \gets cache\_step + 1$
\EndIf
\State Construct draft tree $DT$ using $(TT, d, k)$
\State Verify $DT$ with $M_t$
\State Compute reward $r(s_t, a_t)$ using Eq. \ref{eq:reward_computation}
\State Store trajectory and update policy
\State Reset $cache\_step$ if $cache\_step \geq N$
\end{algorithmic}
\end{algorithm}

\begin{table*}[htbp!]
\centering
\begin{tabular}{llccccc|c|c}
\toprule
\textbf{Backbone} & \textbf{Method} & \textbf{MT-Bench} & \textbf{HumanEval} & \textbf{GSM8K} & \textbf{Alpaca} & \textbf{CNN/DM} & \textbf{Mean} &\textbf{p-value} \\
\midrule
\multirow{4}{*}{LLaMA 3.1-8B}
    & Medusa        & 2.07$\times$ & 2.50$\times$ & 2.23$\times$ & 2.08$\times$ & 1.71$\times$ & 2.12$\times$ & -- \\
    & Hydra         & 2.88$\times$ & 3.28$\times$ & 2.93$\times$ & 2.86$\times$ & 2.05$\times$ & 2.80$\times$ & -- \\
    & EAGLE-3        & 3.39$\times$ & 3.65$\times$ & 3.52$\times$ & 3.67$\times$ & \textbf{2.96$\times$} & 3.44$\times$ & -- \\
    & \textbf{Re-SpS} & \textbf{3.43$\times$} & \textbf{3.89$\times$} & \textbf{3.62$\times$} & \textbf{3.90$\times$} & 2.87$\times$ & \textbf{3.54$\times$} & $<10^{-4}$ \\
\midrule
\multirow{4}{*}{Vicuna-13B}
    & Medusa        & 2.07$\times$ & 2.50$\times$ & 2.23$\times$ & 2.08$\times$ & 1.71$\times$ & 2.12$\times$ & -- \\
    & Hydra         & 2.88$\times$ & 3.28$\times$ & 2.93$\times$ & 2.86$\times$ & 2.05$\times$ & 2.80$\times$ & -- \\
    & EAGLE-3        & 3.75$\times$ & 4.28$\times$ & 3.85$\times$ & 3.76$\times$ & \textbf{3.35$\times$} & 3.80$\times$ & -- \\
    & \textbf{Re-SpS} & \textbf{3.76$\times$} & \textbf{4.64$\times$} & \textbf{3.99$\times$} & \textbf{3.99$\times$} & 3.24$\times$ & \textbf{3.92$\times$} & $<10^{-9}$ \\
\midrule
\multirow{2}{*}{LLaMA 3.3-70B}
    & EAGLE-3        & 4.35$\times$ & 4.87$\times$ & 4.74$\times$ & 4.77$\times$ & \textbf{4.09$\times$} & 4.46$\times$ & -- \\
    & \textbf{Re-SpS} & \textbf{4.47$\times$} & \textbf{5.45$\times$} & \textbf{5.13$\times$} & \textbf{5.34$\times$} & 4.03$\times$ & \textbf{4.88$\times$} & $<10^{-29}$ \\
\bottomrule
\end{tabular}
\caption{Speedup ratios (accepted tokens per second; higher is better) across five benchmarks for LLaMA 3.1-8B, Vicuna-13B, and LLaMA 3.3-70B, comparing Re-SpS with EAGLE-3, Medusa, and Hydra. The p-values are derived from the Wilcoxon signed-rank test, indicating a highly statistically significant difference between the speeds of Re-SpS and EAGLE-3 in all cases. All results are reported at temperature 0. Re-SpS and EAGLE-3 are tested on our machine, while Medusa and Hydra results are from the EAGLE-2~\&~3 papers~\cite{li2024eagle2,li2025eagle3}.}
\label{tab:main-results}
\end{table*}

\section{Experiments}
\label{sec:experiments}

We evaluate Re-SpS against SOTA SpS method, EAGLE-3, across multiple benchmarks and model configurations. Our experimental setup follows established protocols while introducing comprehensive ablation studies to validate our technical contributions.

\subsection{Experimental Setup}

\paragraph{Models and Hardware}
Experiments use three LLM backbones: LLaMA 3.1-8B, Vicuna-13B, and LLaMA 3.3-70B, with their pretrained EAGLE-3 draft models. LLaMA 3.1-8B and Vicuna-13B experiments are conducted on a single NVIDIA A40 GPU (40GB), while LLaMA 3.3-70B requires 4 NVIDIA H100 GPUs (80GB each).

\paragraph{Experimental Setup}
The RL policy is trained using PPO with maximum entropy regularization (entropy coefficient $\beta_H = 0.1$) for enhanced exploration. Training is conducted on a diverse subset of ShareGPT \cite{wangsharegpt} and UltraChat200K \cite{ding2023ultra} datasets containing 4,000 questions across multiple domains. In terms of model architecture, we use a two-layer MLP with 128 hidden units for both the actor and critic networks. 
The detailed experimental settings are in Appendix~A.

\paragraph{Evaluation Benchmarks}
Following EAGLE-3, we evaluate our policy on five common tasks, using the same weights for all tasks without fine-tuning on the respective tasks, 

For multi-turn conversation, code generation, mathematical reasoning, instruction following, and summarization we chose the MT-bench \cite{zheng2023judging}, HumanEval \cite{chen2021evaluating}, GSM8K \cite{cobbe2021training}, Alpaca \cite{alpaca}, and CNN/DailyMail \cite{nallapati2016abstractive} datasets.

\subsection{Main Results}

Table~\ref{tab:main-results} presents the comprehensive performance comparison across all benchmarks. Re-SpS achieves consistent speedup improvements over EAGLE-3 baselines:
\begin{itemize}
    \item \textbf{LLaMA 3.1-8B}: Average speedup of $1.03\times$ over EAGLE-3, with notable improvements on HumanEval ($1.07\times$) and Alpaca ($1.07\times$).
    \item \textbf{Vicuna-13B}: Average speedup of $1.04\times$ over EAGLE-3, with strongest gains on HumanEval ($1.09\times$) and Alpaca ($1.08\times$).
    \item \textbf{LLaMA 3.3-70B}: Average speedup of $1.06\times$ over EAGLE-3, with significant improvements on HumanEval ($1.12\times$) and Alpaca ($1.12\times$).
\end{itemize}

The results demonstrate that dynamic hyperparameter adaptation provides measurable efficiency gains across diverse task domains while maintaining exact output fidelity (byte-for-byte identical to greedy decoding). Statistical significance is assessed with the paired Wilcoxon signed-rank test, with tiny $p$-values showing strong evidence that the difference in inference speed is not noise.

\paragraph{CNN/DailyMail Performance Note}
The slight performance degradation on CNN/DailyMail ($0.98\times$ and $0.97\times$ respectively) results from a necessary evaluation setup modification. To prevent KV cache overflow errors with longer documents, we increased the maximum sequence length from 2048 to 2200 tokens for Re-SpS evaluation, while baselines used the standard 2048-token limit. This difference creates additional computational overhead not present in baseline evaluations.

\subsection{Ablation Study}

\begin{table*}[h]
\centering
\begin{tabular}{@{}llcc@{}}
\toprule
\textbf{Model} & \textbf{Policy Configuration} & \textbf{Avg. Speedup vs. EAGLE-3} & \textbf{Unique Actions} \\
\midrule
\multirow{4}{*}{LLaMA 3.1-8B} 
& Standard PPO + Text Embedding      & 1.044$\times$ & 3  \\
& Standard PPO + Feature Vector   & \textbf{1.049$\times$} & 5  \\
& Max-Entropy PPO + Text Embedding   & 1.017$\times$ & 8  \\
& Max-Entropy PPO + Feature Vector& 1.025$\times$ & \textbf{18} \\
\midrule
\multirow{4}{*}{Vicuna-13B}
& Standard PPO + Text Embedding      & 1.006$\times$ & 8  \\
& Standard PPO + Feature Vector   & 1.028$\times$ & 3  \\
& Max-Entropy PPO + Text Embedding   & 1.015$\times$ & 14 \\
& Max-Entropy PPO + Feature Vector& \textbf{1.033$\times$} & \textbf{15} \\
\bottomrule
\end{tabular}
\caption{Ablation results for policies with a [128,128] hidden layer architecture for both actor and critic networks.}
\label{tab:ablation_128}
\end{table*}
To validate the contributions of our framework's key components, we conduct a series of ablation studies. We analyze the effects of the following components:

\noindent\textbf{1) Feature Representation}: We compare two state representations. The first is a \textit{Text Embedding} where the entire context (prompt, question, and previously generated tokens) is encoded using SentenceBERT \cite{reimers-gurevych-2019-sentence} to provide rich semantic information. The second is a \textit{Feature Vector} representation, which uses the draft model's internal hidden states, avoiding the computational overhead of an external encoder.
The results in Table~\ref{tab:ablation_128} show that the Feature Vector representation consistently outperforms the Text Embedding approach across both LLaMA 3.1-8B and Vicuna-13B backbones, achieving speedups of 1.049$\times$ and 1.033$\times$, respectively. This validates our design choice to leverage internal model features for efficient state representation, as it provides sufficient context without incurring additional inference costs.

\begin{figure}[h!]
\centering
\includegraphics[width=1.0\linewidth]{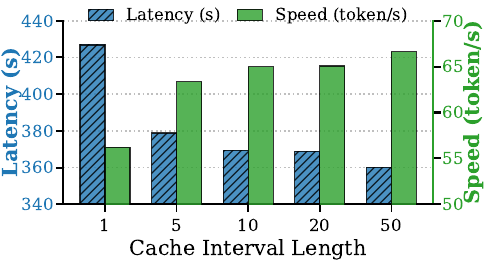}
\caption{Cache interval length vs. inference latency in seconds and generated speed in tokens per second for LLaMA 3.1-8B. The cache interval length is the number of decoding steps over which the RL policy's action is cached and reused. Tested with a fresh SpS RL policy on random 80 questions from the training dataset.}
\label{fig:cache_interval_length}
\end{figure}

\noindent\textbf{2) Cache Interval Length}: We evaluate the impact of the cache interval length, which determines how many decoding steps the RL policy's action is cached and reused. Figure~\ref{fig:cache_interval_length} shows that increasing the cache interval length from 1 to 50 decoding steps leads to a significant reduction in inference time for LLaMA 3.1-8B, while increasing the generated speed (tokens per second). This verifies our assumption that amortizing the cost of policy inference over multiple steps can yield substantial efficiency gains.

\noindent\textbf{3) RL Algorithm}: We compare a \textit{standard PPO} variant against \textit{Max-Entropy PPO}, which includes an entropy regularization term in its objective function. This term encourages the policy to explore a more diverse set of actions rather than converging to a single strategy prematurely. The results in Table~\ref{tab:ablation_128} show that Max-Entropy PPO achieves slightly lower speedups than Standard PPO for LLaMA 3.1-8B (1.025$\times$ vs. 1.049$\times$) but outperforms it for Vicuna-13B (1.033$\times$ vs. 1.028$\times$). However, the Max-Entropy PPO policy exhibits greater action diversity, with 18 unique actions for LLaMA 3.1-8B and 15 for Vicuna-13B, compared to only 5 and 3 for Standard PPO, respectively. This suggests that while Max-Entropy PPO may not always yield the highest speedup, it promotes a more adaptive and robust policy that can better handle diverse contexts.
Other ablations are in Appendix~C.

Overall, these studies show a complex interplay between network capacity, RL algorithm, and the underlying target model. While Max-Entropy PPO consistently promotes action diversity, the optimal configuration for performance depends on the specific model architecture and the capacity of the policy network to leverage either broad exploration or focused, context-driven decision-making.
\section{Conclusion}
We have presented Re-SpS, the first framework to formulate speculative sampling hyperparameter selection as a Markov Decision Process solved through online reinforcement learning. Through evaluation across five diverse benchmarks, Re-SpS consistently outperforms static baselines while preserving exact output fidelity.
Our key contributions demonstrate that dynamic hyperparameter adjustment significantly outperforms static configurations, with our RL agent learning to balance speculative aggression and computational overhead across varying contexts. 
The framework validates that optimal draft tree structures vary substantially across domains, enabling principled optimization of complex trade-offs between the upper limits for the draft tokens, tree depth, and expansion factor.

Future work will extend the framework to other speculative sampling architectures, incorporate sophisticated contextual state representations, and explore multi-objective optimization for throughput, latency, and memory efficiency. The success of RL-based adaptive control suggests broad potential for intelligent optimization across LLM inference acceleration techniques.
\bibliography{aaai2026}
\clearpage
\appendix
\section*{Appendix A: Implementation Details}
\label{app:implementation}
\subsection*{Category Distribution of Questions in Training Dataset}

For training, we use a subset of the ShareGPT \cite{wangsharegpt} and UltraChat200K \cite{ding2023ultra} datasets, comprising 4,000 questions spanning multiple categories. In total, the combined ShareGPT and UltraChat200K datasets contain 266,576 questions distributed across various categories, as detailed in Table \ref{tab:question_distribution}. The largest proportion of questions (41.9\%) belongs to the Writing category, with General (21.9\%) and Reasoning (20.9\%) following closely. The Coding category also represents a notable portion at 7.1\%, while Extraction, Humanities, Math, Roleplay, and other categories account for the remaining percentages.

\begin{table}[htbp!]
\centering
\begin{tabular}{lrr}
\hline
\textbf{Category}    & \textbf{Count} & \textbf{Percentage} \\
\hline
Coding        & 19,058  & 7.1\% \\
Extraction    & 9,111   & 3.4\% \\
General       & 58,490  & 21.9\% \\
Humanities    & 2,625   & 1.0\% \\
Math          & 7,718   & 2.9\% \\
Reasoning     & 55,629  & 20.9\% \\
Roleplay      & 2,157   & 0.8\% \\
Writing       & 111,788 & 41.9\% \\
\hline
\textbf{Total} & 266,576 & 100\% \\
\hline
\end{tabular}
\caption{Category Distribution of Questions.}
\label{tab:question_distribution}
\end{table}

\subsection*{Action Space in Implementation}
During implementation, the action space consists of:
\begin{itemize}
    \item Total tokens ($TT$): $\{32, 48, 64, 80, 96, 128\}$
    \item Tree depth ($d$): $\{3, 4, 5, 6, 7, 8\}$
    \item Top-k expansion factor ($k$): $\{8, 12, 16, 20, 32\}$
\end{itemize}
This yields 180 total combinations, with constraint filtering ensuring $TT \leq k^{d-1}$ to maintain computational feasibility for draft tree construction.

\subsection*{Network Architecture}
We use Stable Baselines 3's Proximal Policy Optimization (PPO) implementation, which includes separate policy and value networks, each with $[128, 128]$ hidden layers.

\subsection*{Hyperparameter Settings}
Table~\ref{tab:hyperparameters} lists the hyperparameters used for training both Standard PPO and Max-Entropy PPO. We use the default values for all hyperparameters for both algorithms, as defined by Stable-Baselines3, with 64 steps instead of 2048 to account for the buffer clearing after every question. entropy coefficient $\beta_H = 0.1$ in Max-Entropy PPO to encourage exploration.

\subsection*{Additional Implementation Details}
When combining and shuffling the ShareGPT and UltraChat200K datasets, the random seed is set to 42 for reproducibility. Training for the Llama 3.1-8B and Vicuna-13B models is performed on a single NVIDIA A40 GPU with 48GB memory, while the Llama 3.3-70B model is trained on four NVIDIA H100 GPUs, each with 80GB memory.

The A40 server is equipped with 61GB of system memory and a 31-core AMD CPU (e.g., AMD EPYC 7313), while the H100 server features 128--256GB of memory and 128--256 core AMD CPUs (e.g., AMD EPYC 9754). All experiments are conducted on Linux operating systems (e.g., Ubuntu 22.04) using Python 3.11.13. For A40 servers, training is managed within a Kubernetes pod. Additional details regarding the setup are provided in the repository's README file.

Training is performed in a single pass over a subset of the ShareGPT and UltraChat200K datasets containing 4,000 questions. Inference is also conducted in a single pass over each complete benchmark dataset, with each benchmark consisting of 80 questions.

\begin{table}[htbp!]
\centering
\begin{tabular}{@{}lcc@{}}
\toprule
Parameter & Standard PPO & Max-Entropy PPO \\
\midrule
Learning Rate & $3 \times 10^{-4}$ & $3 \times 10^{-4}$ \\
PPO Steps & 64 & 64 \\
Batch Size & 32 & 32 \\
PPO Epochs & 4 & 4 \\
Clip Range & 0.2 & 0.2 \\
Gamma & 0.99 & 0.95 \\
GAE Lambda & 0.95 & 0.9 \\
Entropy Coefficient & 0.01 & 0.1 \\
Value Function Coefficient & 0.5 & 0.5 \\
Inference Temperature & 1.0 & 1.5 \\
\bottomrule
\end{tabular}
\caption{Hyperparameter Settings for Standard and Max-Entropy PPO. Max-Entropy PPO is used as the default RL algorithm, with entropy coefficient $\beta_H = 0.1$.}
\label{tab:hyperparameters}
\end{table}


\section*{Appendix B: Algorithm}
\label{app:algorithms}

\subsection*{Re-SpS Draft Tree Construction with Action Caching}
 Algorithm~\ref{alg:resps} represents a single turn in a multi-turn response process, where each question may (e.g., MTBench) or may not (e.g., GSM8K) have multiple turns. Within each turn are a number of intervals, defined by the cache interval $N$, where $N$ is the number of steps in the interval. A step is defined as a single draft/verify run. Configurable hyperparameters $(TT, d,k)$ are adaptively selected by the RL policy every $N$ steps or at the start of a new turn, when applicable.
 The state, action, and cache are reinitialized at the beginning of each turn (lines 1-11). Then, based on the action $a_t$, the hyperparameters $(TT,d,k)$ are defined for the drafting process, which is detailed in lines 12-26.
 The draft tree candidates are verified and accepted tokens are added to the output and used to compute the reward (lines 27-30). Finally, if the cache step is $N$, it resets and updates the PPO policy if the buffer size is satisfied, clearing the buffer either way. At the end of a turn, the final output tokens are returned and used in the context for the next turn.
 
\begin{algorithm}[H]
\caption{Re-SpS Draft Tree Construction with Action Caching}
\label{alg:resps}
\begin{algorithmic}[1]
\Require Target model $M_t$, draft model $M_d$, RL policy $\pi_\theta$, target LLM feature vector $\{h, m, l\}$, cache interval $N$, PPO step parameter $n_{steps}$, and $\text{inference\_mode}\in \{True, False\}$
\Ensure Draft tree $DT$ with adaptive hyperparameters
\State $turn\_done \gets \textbf{False}$
\State Initialize $cache\_step \gets 0$

\State $(TT,d,k) \gets \textbf{None}$
\State Initialize $output\_tokens \gets \emptyset$ \Comment{To store accepted tokens}
\While{\textbf{not $turn\_done$}}

\If{$cache\_step = 0$ \textbf{or} $(TT, d, k)$ are None}
\State Extract state $s_t \gets \text{concat}(h, m, l)$
    \State $a_t = (TT, d, k) \gets \pi_\theta(s_t)$
\Else
    \State $a_t=$ cached $(TT, d, k)$

\EndIf
\State Initialize draft tree $DT$ with root node from $s_t$
\State $curr\_depth \gets 0$
\State $token\_count \gets 1$
\While{$curr\_depth \leq d$}
    \State Choose nodes in layer $l_{curr\_depth}$ with highest global acceptance probabilities $V_i$ as [$n_1, \cdots, n_k$]
        \For {$n \in [n_1, \cdots, n_k]$}
            \State Select highest $V_i$ children nodes $[c_1,\cdots,c_k]$
        \EndFor
    \State Expand to form layer $l_{curr\_depth+1}$ of $DT$ with $[c_1,\cdots, c_k]$
    \State $curr_\_depth \gets curr\_depth+1$
    \State $token\_count \gets token\_count + $(number of new nodes)
    \If{$token\_count \geq TT$ \textbf{or} no viable candidates}
        \State \textbf{break}
    \EndIf
\EndWhile
\State Rerank and verify as in EAGLE-2~\&~3 \cite{li2024eagle2,li2025eagle3}, get accepted tokens 
\State Add accepted tokens to $output\_tokens$
\If {$\text{inference\_mode} = \text{True}$}
\State Compute reward $r(s_t, a_t)$ based on accepted tokens and elapsed time
\State Store trajectory $(s_t, a_t, r_t, cache\_step)$ in experience buffer

\EndIf

\If{\textbf{EOS} $\in \text{accepted tokens}$ \textbf{or} $|output\_tokens| \ge T_{\max}$} 
    \State $turn\_done \gets \textbf{True}$    
   \EndIf
   \If{$cache\_step = N$}                                        \Comment{interval finished}
        \State $cache\_step \gets 0$
   \State $cache\_step \gets cache\_step+1$
    \If{buffer size $\ge n_{steps}$}              
        \State Update PPO policy with batch
        \State Clear buffer
    \EndIf
\EndIf
\EndWhile
\State Return $output\_tokens$
\end{algorithmic}
\end{algorithm}

\begin{table*}[htbp!]
\centering
\begin{tabular}{@{}llcc@{}}
\toprule
\textbf{Model} & \textbf{Policy Configuration} & \textbf{Avg. Speedup vs. EAGLE-3} & \textbf{Unique Actions} \\
\midrule
\multirow{4}{*}{LLaMA 3.1-8B} 
& Standard PPO + Text Embedding      & \textbf{1.033$\times$} & 2  \\
& Standard PPO + Feature Vector   & 1.027$\times$ & 3  \\
& Max-Entropy PPO + Text Embedding   & 1.028$\times$ & 8  \\
& Max-Entropy PPO + Feature Vector& 0.981$\times$ & \textbf{12} \\
\midrule
\multirow{4}{*}{Vicuna-13B}
& Standard PPO + Text Embedding      & \textbf{1.056$\times$} & 1  \\
& Standard PPO + Feature Vector   & 1.040$\times$ & 2  \\
& Max-Entropy PPO + Text Embedding   & 1.021$\times$ & 7  \\
& Max-Entropy PPO + Feature Vector& 1.002$\times$ & \textbf{8}  \\
\bottomrule
\end{tabular}
\caption{Ablation results for policies with a [64,64] hidden layer architecture for both actor and critic networks.}
\vspace{-4.3mm}
\label{tab:ablation_64}
\end{table*}
\section*{Appendix C: Additional Ablation Studies}
\label{app:ablations}
\noindent\textbf{Policy Network Capacity}: The policy actor and critic networks in our default implementation are MLPs with two hidden layers of [128,128] units. In this section, we explore the impact of network capacity on performance by evaluating configurations with [64,64] hidden layers.
With [64,64] networks (Table \ref{tab:ablation_64}), Standard PPO with Text Embedding yields the highest speedups ($1.033\times$ for LLaMA 3.1-8B and $1.056\times$ for Vicuna-13B). For Vicuna-13B, this configuration is the top performer, achieving its result by converging to a single, highly effective action. This indicates that with a smaller network, a decisive, context-guided policy is most effective. In contrast, the Max-Entropy policies are more exploratory, using 7-12 unique actions, but this diversity does not translate into superior performance at this network size.

Why does this happen? The feature vector representation from EAGLE-3~\cite{li2025eagle3} has dimension $[3, 4096]$, while the text embedding representation has a much smaller dimension of $[384]$. Therefore, compared to the default [128,128] network, the [64,64] network hardly has enough capacity to learn the complex feature vector representation of $[3, 4096]$, leading to suboptimal performance. In contrast, the text embedding representation is much smaller, allowing the [64,64] network to learn a more effective policy.
\section*{Appendix D: Additional Analysis}

\subsection{Acceptance Rate and Acceptance Length}

We further analyze the acceptance behavior of Re-SpS compared to the EAGLE-3 baseline. Table \ref{tab:acceptance_metrics} details the acceptance rate and acceptance length across five standard benchmarks. These experiments were conducted using the \text{LLaMA 3.1-8B} backbone.

As shown in the table, Re-SpS achieves a higher average acceptance rate compared to EAGLE-3 ($0.2335$ vs $0.2072$). Re-SpS achieves a lower average of $43.44$ tokens, compared to $49.11$ for EAGLE-3. Although Re-SpS exhibits a shorter average acceptance length, the higher acceptance rate implies that Re-SpS effectively optimizes the draft structure to maximize verification throughput rather than just maximizing raw draft length. During the training phase, we observed an average acceptance rate of $0.2113 \pm 0.0472$ and an average acceptance length of $58.88 \pm 23.61$ tokens.

\begin{table*}[t]
    \centering
    \begin{tabular}{lcccccc}
        \toprule
        \textbf{Metric} & \textbf{MT-Bench} & \textbf{HumanEval} & \textbf{GSM8K} & \textbf{Alpaca} & \textbf{CNN/DM} & \textbf{Mean} \\
        \midrule
        \multicolumn{7}{l}{\textit{Acceptance Rate}} \\
        EAGLE-3 & $0.2032 \pm 0.0300$ & $0.2214 \pm 0.0207$ & $0.2101 \pm 0.0194$ & $0.2281 \pm 0.0325$ & $0.1732 \pm 0.0159$ & $0.2072$ \\
        Re-SpS & $0.2029 \pm 0.0647$ & $0.2419 \pm 0.0565$ & $0.2519 \pm 0.1168$ & $0.2315 \pm 0.0519$ & $0.2395 \pm 0.0821$ & $0.2335$ \\
        \midrule
        \multicolumn{7}{l}{\textit{Acceptance Length}} \\
        EAGLE-3 & $60.44 \pm 29.04$ & $60.15 \pm 17.29$ & $34.38 \pm 10.84$ & $36.75 \pm 27.62$ & $53.83 \pm 18.83$ & $49.11$ \\
        Re-SpS & $53.49 \pm 25.97$ & $50.25 \pm 16.31$ & $29.95 \pm 11.16$ & $30.82 \pm 24.41$ & $52.67 \pm 19.02$ & $43.44$ \\
        \bottomrule
    \end{tabular}
    \caption{Detailed comparison of Acceptance Rate and Acceptance Length between EAGLE-3 and Re-SpS on the LLaMA 3.1-8B model. Values are reported as Mean $\pm$ Standard Deviation. Re-SpS demonstrates competitive acceptance rates while maintaining substantial acceptance lengths across diverse tasks.}
    \vspace{-4.3mm}
    \label{tab:acceptance_metrics}
\end{table*}

\subsection{System Profiling and Overhead Analysis}

To understand the computational cost of the proposed method, we performed a detailed profiling breakdown of the system overhead. The profiling was conducted on 80 examples during the training phase; consequently, the reported times include the overhead from training updates. The analysis focuses on identifying the time consumed by specific components during this process.
We categorize the overhead into four mutually exclusive components to provide a clear view of the computational distribution:
\begin{itemize}
    \item \textbf{Drafting}: The core process of generating candidate tokens.
    \item \textbf{Tree Structure Management}: A composite category that includes draft tree construction, initialization, updates, and input updates. Notably, the \textit{input update} operation handles the synchronization of the model's state for the next autoregressive step following verification. This involves updating the Key-Value (KV) cache to discard rejected tokens, adjusting attention masks, and preparing position IDs.
    \item \textbf{Verification}: The target model's verification step.
    \item \textbf{RL Policy Prediction}: The time taken by the reinforcement learning agent to infer the optimal tree parameters.
\end{itemize}

Table \ref{tab:components} details the breakdown of components, showing only leaf components with meaningful execution time.

\begin{table}[htbp!]
    \centering
    \begin{tabular}{lc}
        \toprule
        \textbf{Component} & \textbf{Time (\%)} \\
        \midrule
        drafting\_process & 9.9 \\
        tree\_update & 3.5 \\
        input\_update & 3.5 \\
        tree\_construction & 0.6 \\
        tree\_initialization & 0.6 \\
        verification\_process & 0.1 \\
        post\_processing & 0.0 \\
        tokenization & 0.0 \\
        rl\_policy\_prediction & 0.0 \\
        hidden\_states\_extraction & 0.0 \\
        rl\_policy\_update & 0.0 \\
        \midrule
        Total meaningful time & 18.2 \\
        Container/organizational overhead & 5.9 \\
        Unaccounted time (model inference, etc.) & 75.9 \\
        \textbf{TOTAL TIME} & \textbf{100.0} \\
        \bottomrule
    \end{tabular}
    \caption{Components Breakdown showing leaf components with meaningful execution time. The majority of time is spent in model inference (unaccounted) and drafting.}
    \vspace{-4.3mm}
    \label{tab:components}
\end{table}

\subsubsection{Consolidated Component Analysis}

To address the overlap in inclusive components, we consolidated specific operations into four mutually exclusive categories. It is important to note that this breakdown includes only specific components contributing to the \text{Re-SpS framework overhead}, rather than an exhaustive list of all system operations. As shown in Table \ref{tab:consolidated}, the vast majority of the measured overhead (99.7\%) is attributed to \text{Drafting} (54.6\%) and \text{Tree Structure Management} (45.1\%). Notably, the RL Policy Prediction overhead is negligible ($0.02\%$), confirming the efficiency of our action caching and feature reuse strategies.

\begin{table}[htbp!]
    \centering
    \begin{tabular}{lc}
        \toprule
        \textbf{Component} & \textbf{Time (\%)} \\
        \midrule
        Drafting & 54.61\% \\
        Tree Structure Management & 45.08\% \\
        Verification & 0.28\% \\
        RL Policy Prediction & 0.02\% \\
        \midrule
        \textbf{Total} & \textbf{100.00\%} \\
        \bottomrule
    \end{tabular}
    \caption{Consolidated Component Breakdown for LLaMA 3.1-8B. Tree Structure Management aggregates draft tree construction, initialization, updates, and input updates.}
    \label{tab:consolidated}
\end{table}
\vspace{-4.3mm}
\subsection{Training Efficiency and PPO Variant Analysis}

\paragraph{Training Cost}
We report the wall-clock training time for the RL policy across different model sizes. The training was conducted on a single node with NVIDIA GPUs. The total training times are:
\begin{itemize}
    \item \textbf{LLaMA 3.3-8B:} 8.65 hours
    \item \textbf{Vicuna-13B:} 10.13 hours
    \item \textbf{LLaMA 3.1-70B:} 11.05 hours
\end{itemize}
\end{document}